\UseRawInputEncoding

\documentclass[letterpaper, 10 pt, conference]{ieeeconf}  % Comment this line out if you need a4paper

\IEEEoverridecommandlockouts                              % This command is only needed if 
\usepackage[table]{xcolor}
\usepackage{float}
\usepackage{amsmath}
\usepackage{threeparttable}
\usepackage{multirow}
\usepackage{graphicx}
\usepackage{amssymb}
\usepackage{booktabs}
\usepackage{hyperref}
\definecolor{ccr}{RGB}{255,0,0}
\hypersetup{
	hypertex=true,
	colorlinks=true,
	linkcolor=ccr,
	anchorcolor=ccr,
	citecolor=ccr
}

\title{FGS-SLAM: Fourier-based Gaussian Splatting for Real-time SLAM with Sparse and Dense Map Fusion
}

\author{Yansong Xu$^{1,2}$, Junlin Li$^{1}$, Wei Zhang$^{1}$, Siyu Chen$^{1,2}$, Shengyong Zhang$^{1}$, Yuquan Leng$^{3*}$, Weijia Zhou$^{1*}$% <-this % stops a space
\thanks{*Corresponding author.}% <-this % stops a space
\thanks{Yansong Xu and Siyu Chen are with $^{1}$ State Key Laboratory of Robotics, Shenyang Institute of Automation, Chinese Academy of Sciences, Shenyang 110016, China, and also with  $^{2}$ University of Chinese Academy of Sciences, Beijing 100049, China.
        {\tt\small \{xuyansong21, chensiyu23\}@mails.ucas.ac.cn}}%
\thanks{Junlin Li, Wei Zhang, Weijia Zhou and Shengyong Zhang are with $^{1}$ State Key Laboratory of Robotics, Shenyang Institute of Automation, Chinese Academy of Sciences, Shenyang 110016, China.  
        {\tt\small\{lijunlin, zhangwei, zwj, zhangshengyong\}@sia.cn}}%
\thanks{Yuquan Leng is with $^{3}$ Department of Mechanical and Energy Engineering, Southern University of Science and Technology, Shenzhen 518055, China.  
	{\tt\small lengyq@sustech.edu.cn}}%
}

\begin{document}

\maketitle
\thispagestyle{empty}
\pagestyle{empty}

%%%%%%%%%%%%%%%%%%%%%%%%%%%%%%%%%%%%%%%%%%%%%%%%%%%%%%%%%%%%%%%%%%%%%%%%%%%%%%%%
\begin{abstract}

3D gaussian splatting has advanced simultaneous localization and mapping (SLAM) technology by enabling real-time positioning and the construction of high-fidelity maps. However, the uncertainty in gaussian position and initialization parameters introduces challenges, often requiring extensive iterative convergence and resulting in redundant or insufficient gaussian representations. To address this, we introduce a novel adaptive densification method based on Fourier frequency domain analysis to establish gaussian priors for rapid convergence. Additionally, we propose constructing independent and unified sparse and dense maps, where a sparse map supports efficient tracking via Generalized Iterative Closest Point (GICP) and a dense map creates high-fidelity visual representations. This is the first SLAM system leveraging frequency domain analysis to achieve high-quality gaussian mapping in real-time. Experimental results demonstrate an average frame rate of 36 FPS on Replica and TUM RGB-D datasets, achieving competitive accuracy in both localization and mapping. The source code is publicly available at \url{https://github.com/3DV-Coder/FGS-SLAM}.

\end{abstract}

%%%%%%%%%%%%%%%%%%%%%%%%%%%%%%%%%%%%%%%%%%%%%%%%%%%%%%%%%%%%%%%%%%%%%%%%%%%%%%%%

\begin{figure}[!t]
	\centering
	
	\includegraphics[width=0.45\textwidth]{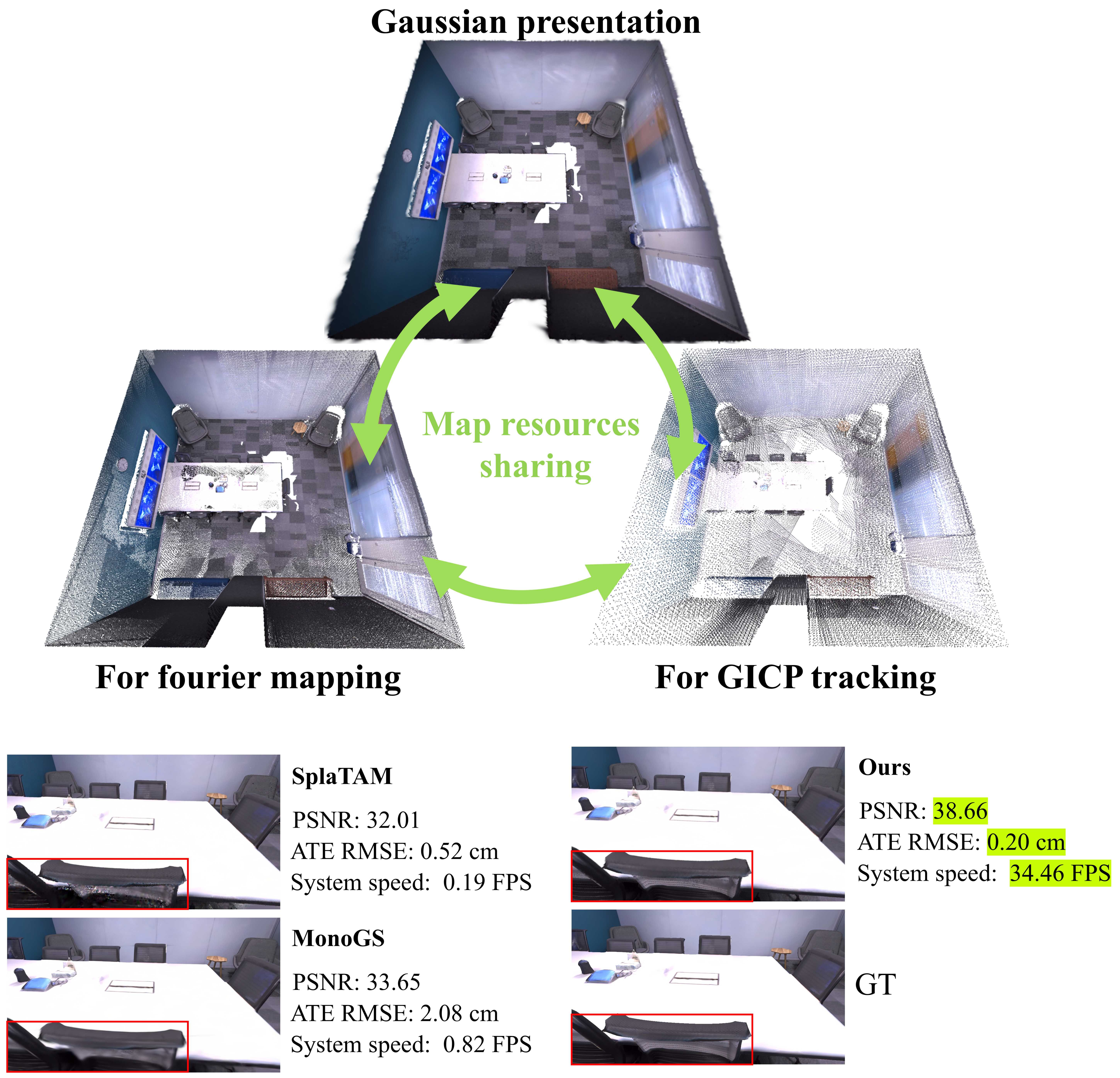}
	\caption{FGS-SLAM adopts a map-sharing mechanism, jointly maintaining both a 3D gaussian dense map and a sparse map. The gaussian map provides excellent rendering performance from new viewpoints. The system operates at a speed at least one order of magnitude faster than other methods, achieving real-time frame rates, while ensuring accurate localization and high-fidelity map reconstruction quality.}
	\label{fig1}
\end{figure}

\section{Introduction}

With the rapid development of fields such as robotics \cite{li2024robust}, augmented reality (AR), and drones, there is an increasing demand for efficient and accurate 3D environmental perception. As a result, the importance of Simultaneous Localization and Mapping (SLAM) technology in these applications has grown significantly. The main challenge of SLAM is achieving both autonomous localization and 3D map construction in unknown environments. However, traditional SLAM methods still face significant challenges in balancing real-time performance and scene accuracy.

In recent years, sparse and dense SLAM methods have emerged as key approaches. Sparse SLAM methods are computationally efficient and offer advantages in real-time performance \cite{mur2017orb, campos2021orb}. However, they represent the scene with fewer features, limiting the ability to reconstruct detailed environments. Dense SLAM methods \cite{dai2017bundlefusion, whelan2015elasticfusion, Whelan2012KintinuousSE, wang2024fi}, on the other hand, create fine-grained environmental representations using high-density point clouds and voxel models. However, their high computational complexity makes them less suitable for real-time applications. To overcome these limitations, researchers have explored neural network-based implicit dense map construction, such as Neural Radiance Fields (NeRF) SLAM. These methods use neural radiance fields to enhance scene detail \cite{zhu2022nice, wang2023co, johari2023eslam, yang2022vox}. However, the high computational cost of volumetric rendering makes it difficult to achieve real-time performance, and the maps lack interpretability.

3D gaussian splatting (3DGS) is an explicit dense mapping method that represents the environment with gaussian distributions \cite{kerbl20233d, sun2024mm3dgs}. This approach offers high rendering speed and better interpretability. SLAM systems based on 3DGS have significantly improved pose estimation efficiency and map quality, making it an important direction in explicit dense SLAM \cite{keetha2024splatam, matsuki2024gaussian, sun2024mm3dgs}. However, current 3DGS methods often struggle with the uncertainty in gaussian point initialization, leading to redundancy or under representation, which impacts mapping efficiency and accuracy. While existing 3DGS-based methods predominantly rely on spatial domain information for map construction, FreGS \cite{zhang2024fregs} mitigates over-reconstruction by introducing frequency regularization to supervise the gaussian optimization process in the spectral domain. In contrast to FreGS, our approach focuses on optimizing gaussian initialization and distribution strategies directly in the frequency domain."

To address these challenges, this paper introduces FGS-SLAM, a method that adapts gaussian densification based on frequency domain analysis. For the first time, we use Fourier domain analysis for gaussian initialization, reducing redundant gaussian points and accelerating parameter convergence. Additionally, we propose an independent yet unified approach to constructing sparse and dense maps. The sparse map is used for efficient camera tracking, while the dense map enables high-fidelity scene representation. This method enables a SLAM system that balances real-time performance with high accuracy.

The main contributions of this paper are as follows:
\begin{itemize}
	\item \textbf{Gaussian Initialization Strategy Guided by Frequency Domain:} We propose a novel frequency-domain analysis-based gaussian initialization method, which accelerates the convergence of gaussian parameters and reduces redundant points, thus enhancing the efficiency of dense map construction.
	\item \textbf{Independent Unified Framework for Sparse and Dense Maps:} We achieved efficient localization with sparse maps and high-fidelity reconstruction with dense maps, effectively balancing the real-time performance and detailed expression capabilities of the SLAM system.
	\item \textbf{Adaptive gaussian Density Distribution Strategy:} By leveraging frequency domain analysis, we adaptively assign gaussian density and radius to different regions of the scene, effectively reducing computational complexity while ensuring scene accuracy.
\end{itemize}

\section{Related Works}
\subsection{Sparse map SLAM} 
Sparse map SLAM typically focuses on real-time performance and computational efficiency by selecting a limited number of key feature points for camera tracking and pose estimation. Representative methods include ORB-SLAM2\cite{mur2017orb} and ORB-SLAM3 \cite{campos2021orb}. These methods reduce computational cost by extracting feature points to form sparse maps but are limited in their ability to express scene details. The Generalized Iterative Closest Point (GICP) algorithm \cite{segal2009generalized} is a widely used sparse SLAM tracking method that relies on point cloud matching for accurate pose estimation, making it suitable for sparse map construction. Despite the broad application of ICP-based algorithms in sparse map SLAM, their accuracy is highly dependent on the density and quality of the map, which makes it challenging to achieve precise scene reconstruction in complex environments. The FGS-SLAM method proposed in this paper combines the efficiency of sparse maps with the high-fidelity scene representation of dense maps. By addressing the redundancy issue with frequency-domain-guided gaussian initialization, FGS-SLAM achieves a SLAM system that balances both real-time performance and precision.

\subsection{Dense map SLAM}
Dense map SLAM, on the other hand, provides a complete representation of the environment through high-density point clouds or voxel modeling \cite{yang2022vox, hourdakis2021roboslam}. Methods like ElasticFusion \cite{whelan2015elasticfusion} and KinectFusion \cite{newcombe2011kinectfusion} construct dense maps using depth sensors, enabling high-quality scene reconstruction. However, these methods suffer from poor real-time performance and high computational resource requirements. While dense SLAM has made significant advancements in scene detail representation, it still faces challenges in real-time applications. This paper addresses these challenges by introducing an adaptive gaussian density distribution strategy, achieving efficient dense map construction while maintaining both high-fidelity scene representation and the real-time demands of the SLAM system.

\subsection{NeRF SLAM} 
The introduction of Neural Radiance Fields has allowed SLAM systems to achieve implicit dense scene representations, thereby enhancing the level of detail in 3D map reconstruction. iMAP \cite{sucar2021imap} was the first real-time NeRF SLAM system that jointly optimized the 3D scene and camera poses through implicit neural networks. NICE-SLAM \cite{zhu2022nice} further employed feature grids and multi-resolution strategies to accelerate the optimization of dense SLAM scenes while retaining NeRF's fine-grained representation. Co-SLAM \cite{wang2023co} improved NeRF's performance in both detail expression and efficient optimization by using a multi-resolution hash grid structure. While NeRF-based SLAM methods enable high-quality scene reconstruction, the computational burden of volume rendering and ray tracing leads to poor real-time performance, and the implicit representation reduces the interpretability of the map.

\subsection{3DGS SLAM} 
3D gaussian splatting method was introduced to enable explicit dense map construction. 3DGS uses gaussian points to directly represent the scene, offering faster rendering speeds and better interpretability. Methods such as SplaTAM \cite{keetha2024splatam} and GS-SLAM \cite{yan2024gs} leverage the fast rendering capabilities of 3DGS to construct high-fidelity dense maps, significantly improving efficiency. However, the uncertainty in the initialization of gaussian points in 3DGS often leads to redundant or under-expressed gaussian points, which impacts mapping efficiency and accuracy. The FGS-SLAM method proposed in this paper improves on 3DGS by using frequency-domain information for gaussian initialization, allowing the parameters to converge more quickly, reducing redundancy, and improving the quality of the dense map, ultimately enhancing both localization and mapping accuracy in SLAM systems.

\section{Method}
\begin{figure*}[!t]
	\centering
	\includegraphics[width=0.85\textwidth]{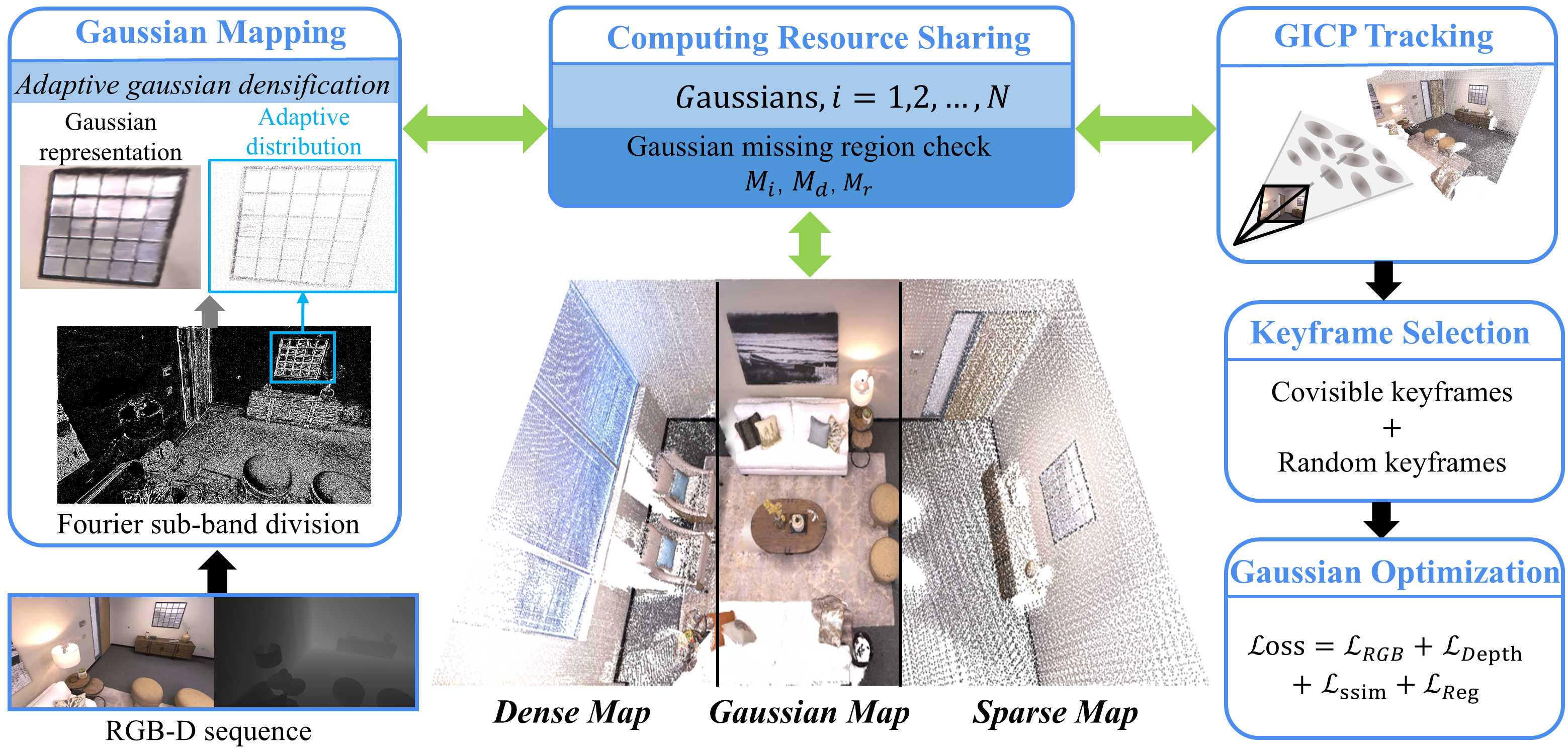}
	\caption{System Overview. The proposed method uses RGB-D data as input to the system. Mapping: The spatial domain is transformed into the frequency domain through Fourier transforms. New gaussians are adaptively initialized based on high and low frequency regions, thereby constructing a gaussian dense map. Resource Sharing: The system simultaneously constructs both sparse and dense maps, with map points stored using gaussian attributes. Gaussian attributes and the mask of missing gaussian regions are shared between the maps. Tracking: GICP performs rapid registration using the 3D gaussian point cloud of the sparse map, and supplements the gaussian points in the sparse map. The system selects co-visibility and random keyframes, and jointly optimizes the gaussian map through gaussian rasterization rendering.}
	\label{fig2}
\end{figure*}

Fig. \ref{fig2} illustrates the system framework. Both the dense and sparse maps are composed of 3D gaussians. The gaussian set possesses the following properties: $G\{\mu_{i}, S_{i}, R_{i}, \alpha_{i}, c_{i}\}, (i=1, \ldots, N)$. Each gaussian consists of a position $\mu_{i}$, scale $S_{i}$, rotation $R_{i}$, opacity $\alpha_{i}$, and color $\mathrm{c}_{i}$. The relationship between the scale $S$ and covariance $C$ is given by $C = RSS^\mathrm{T}R^\mathrm{T}$. The 3D gaussian splatting rendering process is based on alpha blending, which achieves the 2D projection. In this paper, we describe the rendering process using the following equation:
\begin{equation}
	\label{eq1}
	\mathcal{F}_{p}=\sum_{i=1}^{n}\mathbf{\gamma}_{i}\alpha_i\prod_{j=1}^{i-1}(1-\alpha_{j}),
\end{equation}
where $\mathcal{F}{p}$ represents the rendered value of the pixel formed by the combination of $n$ gaussians splats, and $\gamma_{i}$ denotes the contribution coefficient of the $i$-th gaussian to the pixel rendering. This equation can be interpreted differently based on the type of rendering:
1) Color Rendering: Here, $\gamma_{i}$ indicates the color of each gaussian $c_{i}$, with the equation capturing the cumulative color contribution.
2) Depth Rendering: For depth rendering, $\gamma_{i}$ denotes the depth of each gaussian $d_{i}$, and the process accumulates depth values across gaussians.
3) Opacity Rendering: By setting $\gamma_{i}=1$ to represent opacity, the equation simplifies to model the accumulation of opacity, which is essential for computing visibility under a specific viewpoint.

Gaussian distribution assumptions in GICP and 3DGS share a common foundation, utilizing a shared gaussian set $G\{\mu_{i},S_{i}\},(i=1,\ldots,N)$. This shared representation enables efficient tracking via GICP, which estimates poses based on sparse gaussian maps, and high-quality mapping via 3DGS, which optimizes and updates the map using these poses. The integration of GICP and 3DGS leverages 3D gaussian representations to achieve fast and accurate SLAM performance.

Functionally, the sparse and dense maps operate independently: the sparse map facilitates efficient camera tracking, while the dense map supports map reconstruction. Both maps are unified through shared gaussian attributes and joint optimization. Consistency in gaussian attribute optimization is maintained via joint rendering of points from both maps. The dense map adaptively initializes new gaussians, shares missing-region masks with the sparse map, and filters tracking points, ensuring resource sharing and optimization coherence.

\subsection{Mapping}
\textbf{Adaptive Gaussian Densification.} The color change gradient corresponds to the frequency. The scene is composed of color change gradients at different frequencies. We observe that regions with small color change gradients, representing low-frequency areas, are expected to use sparse, large gaussian representations, while dense, small gaussian representations are more suitable for high-frequency areas. To address this, we propose adaptive gaussian densification.  The Fourier transform is used to convert the spatial domain of the image frame $I(x,y)$ into the frequency domain. The Fourier domain representation of the image is expressed as:
\begin{equation}
	\label{eq2}
	F(u,v)=\sum_{x=0}^{W-1}\sum_{y=0}^{H-1}I(x,y)\cdot e^{-j2\pi(\frac{u\cdot x}{W}+\frac{v\cdot y}{H})},
\end{equation}
where $F_c(u,v)$ represents the complex-valued function in the frequency domain at $(u,v)$, and $I(x,y)$ denotes the pixel values in the spatial domain. $W$ and $H$ are the width and height of the image, and $u$ and $v$ are the horizontal and vertical coordinates in the frequency domain. The Fourier centering is defined as:
\begin{equation}
	\label{eq3}
	F_c(u,v)=F(u,v)\cdot (-1)^{u+v}.
\end{equation}

The magnitude spectrum is given by:
\begin{equation}
	\label{eq4}
	\begin{vmatrix}F_c(u,v)\end{vmatrix}=\sqrt{Re(u,v)^2+Im(u,v)^2},
\end{equation}
where $Re(u,v)$ and $Im(u,v)$ denote the real and imaginary parts of the complex-valued function $F_c(u,v)$. Additionally, a gaussian filter is applied to the frequency domain image for frequency separation. The frequency domain transfer function of the gaussian high-pass filter is:
\begin{equation}
	\label{eq5}
	H(u,v)=1-e^{-\frac{D^2(u,v)}{2D^2_0}},
\end{equation}
where $D(u,v)$ is the Euclidean distance from the pixel position $(u,v)$ to the filter center. $D_0$ is the cutoff frequency of the filter, determining the strength of the high-pass filter. 

After applying the gaussian high-pass filter, the frequency domain function $F_h(u,v)$ is:
\begin{equation}
	\label{eq7}
	F_h(u,v) = F_c(u,v)\cdot H(u,v).
\end{equation}

The Fourier inverse transform of the gaussian high-pass filter is also a gaussian function. This means that the inverse Fourier transform (IDFT) of the equation above results in a spatial gaussian filter that avoids ringing effects. This filter sets the low-frequency direct current component to zero, meaning the filtered result depends only on the scene's color gradient changes and is not influenced by the scene's color.

The high-frequency component-dominated image is reconstructed in the spatial domain after inverse Fourier transform from the filtered frequency domain image:
\begin{equation}
	\label{eq8}
	\tilde{I_h}(x,y)= \frac{1}{H\cdot W}\sum_{x=0}^{H-1}\sum_{y=0}^{W-1}F_h(u,v)\cdot e^{j2\pi(u\frac{x}{W}+v\frac{y}{H})}.
\end{equation}

Typically, the energy from low to high frequencies decreases overall. Therefore, the energy values obtained after gaussian high-pass filtering decrease from low to high, and the frequency histogram of the high-frequency image $\tilde{I_h}(x,y)$ presents a unimodal shape. Based on this observation, we use a triangular thresholding method to construct a triangle between the highest peak of the histogram and its endpoints, finding the point furthest from the baseline as the threshold. The high-frequency region is then segmented, and the low-frequency region is obtained by complement. We use equidistant sampling points with varying spacings in different frequency domains as gaussian sampling points, with the sampling interval in high-frequency regions being $m$, and in low-frequency regions being $n$ where $(m < n)$. The resulting high-frequency region position mask is $M_h$. The low-frequency region position mask is $M_l$ after the inversion of $M_h$.

\textbf{Gaussian Missing Region Check Strategy.} Incorporating all gaussians blindly into the map would result in gaussian explosion and redundancy. Therefore, we propose a gaussian missing region check strategy. The gaussian missing region refers to areas that were not observed in previous keyframes or regions where the gaussian map representation is insufficient. Using equation (\ref{eq1}), we perform alpha blending for opacity rendering (equivalent to the accumulation of gaussian opacity). If the final rendering opacity is lower than the threshold, the area is considered insufficient in gaussian representation, resulting in the gaussian expression deficit mask $M_i$ in the current frame. Additionally, considering the presence of foreground and background in the scene, if the background has already been mapped in previous keyframes but the foreground is present in the current frame's viewpoint, simply using opacity rendering cannot detect the missing foreground. Thus, we introduce a depth mask $M_d$ and a color mask $M_c$. The depth mask is generated by comparing the gaussian depth rendering with the current frame's depth. Areas with an abnormally large depth difference are considered foreground. Similarly, the color mask compares the gaussian color rendering with the current frame's color. Areas with significant color differences are treated as foreground. The final gaussian missing region mask $M_m$ is:
\begin{equation}
	\label{eq9}
	M_m = M_i \cup M_d \cup M_c.
\end{equation}

The final regions where gaussians are added are:
\begin{equation}
	\label{eq10}
	\begin{aligned}
		M_h
		&=M_h \cap M_m\\
		M_l&=M_l \cap M_m\\
	\end{aligned}.
\end{equation}

Different gaussian radii are set for high and low-frequency regions:
\begin{equation}
	\label{eq11}
	r =
	\begin{cases} 
		\alpha_h \cdot \frac{ d}{f} & \text{if } G \in M_h \\
		\alpha_l \cdot\frac{d}{f} & \text{if } G \in M_l 
	\end{cases},
\end{equation}
where the ratio of depth $d$ to focal length $f$ represents the gaussian scale based on the 3D gaussian projection onto a 2D image, with a radius of 1 pixel. $\alpha_h$ and $\alpha_l$ represent the gaussian scale factors for high-frequency and low-frequency regions, respectively, where $\alpha_h < \alpha_l$ means the gaussian scale for high-frequency regions is smaller than that for low-frequency regions. Instead of performing gaussian splitting after gaussian initialization, new gaussians are adaptively added in the regions where gaussians are missing.

\textbf{Gaussian Pruning.} The gaussian pruning strategy includes two components. First, gaussian overgrowth in any dimension may cause artifacts in the gaussian map. Thus, we prune excessively large gaussians and use the mapping strategy to re-supplement them. Secondly, gaussians with low opacity contribute weakly to scene representation. To reduce redundancy, we prune gaussians with opacity below the threshold.

\subsection{Tracking}
We construct a sparse gaussian map as the target point cloud for efficient tracking via Generalized Iterative Closest Point (GICP) alignment. To prevent tracking accuracy and speed degradation from excessive point clouds, we apply uniform downsampling on keyframes to build the sparse map in 3D space. The subsampling point of the current frame serves as the source point cloud, while the sparse gaussian map is the target point cloud for ICP tracking.

The sparse map is updated only on tracking keyframes. The gaussian addition strategy is similar to that of the dense map. The source point cloud is filtered through the gaussian missing region mask. It is then added to the sparse map.

Tracking is achieved through the GICP method, which optimizes a transformation matrix $T$ to align the source point cloud $P=\{p_0,p_1,\ldots,p_M\}$ with the target point cloud $Q=\{q_0,q_1,\ldots,q_N\}$. GICP models the local surfaces of the source point $p_i$ and the target point $q_i$ as gaussian distributions: $p_i\sim \mathcal{N}(\hat{p_i},C_p^i)$, $\quad q_i\sim \mathcal{N}(\hat{q_i},C_q^i)$, where $C_p^i$ and $C_q^i$ are the covariance matrices of the local regions of $p_i$ and $q_i$. The registration error between the two point clouds is defined as:
\begin{equation}
	\label{eq12}
	\hat{d}_i=\hat{q}_i-T\hat{p}_i,
\end{equation}

Using the properties of gaussian distributions, the error $d_i$ is derived to follow:
\begin{equation}
	\label{eq13}
	d_i\sim \mathcal{N}(0,C_q^i+TC_p^iT^\mathrm{T}),
\end{equation}

To find the optimal transformation matrix $T$, we maximize the probability distribution of each paired point $\mathbf{p}(d_{i})$ (maximum log-likelihood estimation):
\begin{equation}
	\label{eq14}
	\begin{aligned}
		T
		&=\underset{\mathbf{T}}{\mathrm{argmax}}\prod_{i}\mathbf{p}(d_{i})\\
		&=\underset{\mathbf{T}}{\mathrm{argmax}}\sum_{i}\log(\mathbf{p}(d_{i}))\\
		&=\arg\min_T\sum_id_i^\mathrm{T}\left(C_q^i+TC_p^iT^\mathrm{T}\right)^{-1}d_i
	\end{aligned},
\end{equation}

\subsection{Keyframe Selection}
Tracking keyframes are selected based on the overlap ratio between the observed point cloud in the current frame and the sparse gaussian map. For points within a permissible distance, they are considered overlapping. If the ratio of overlapping points to the total observed point cloud is below a threshold, the current frame is chosen as a tracking keyframe. If the current frame differs from the previous tracking keyframe by ten frames, it is marked as a mapping-only keyframe. Tracking keyframes serve both tracking and mapping purposes, while mapping keyframes are used solely for mapping.

\begin{figure}[!t]
	\centering
	
	\includegraphics[width=0.42\textwidth]{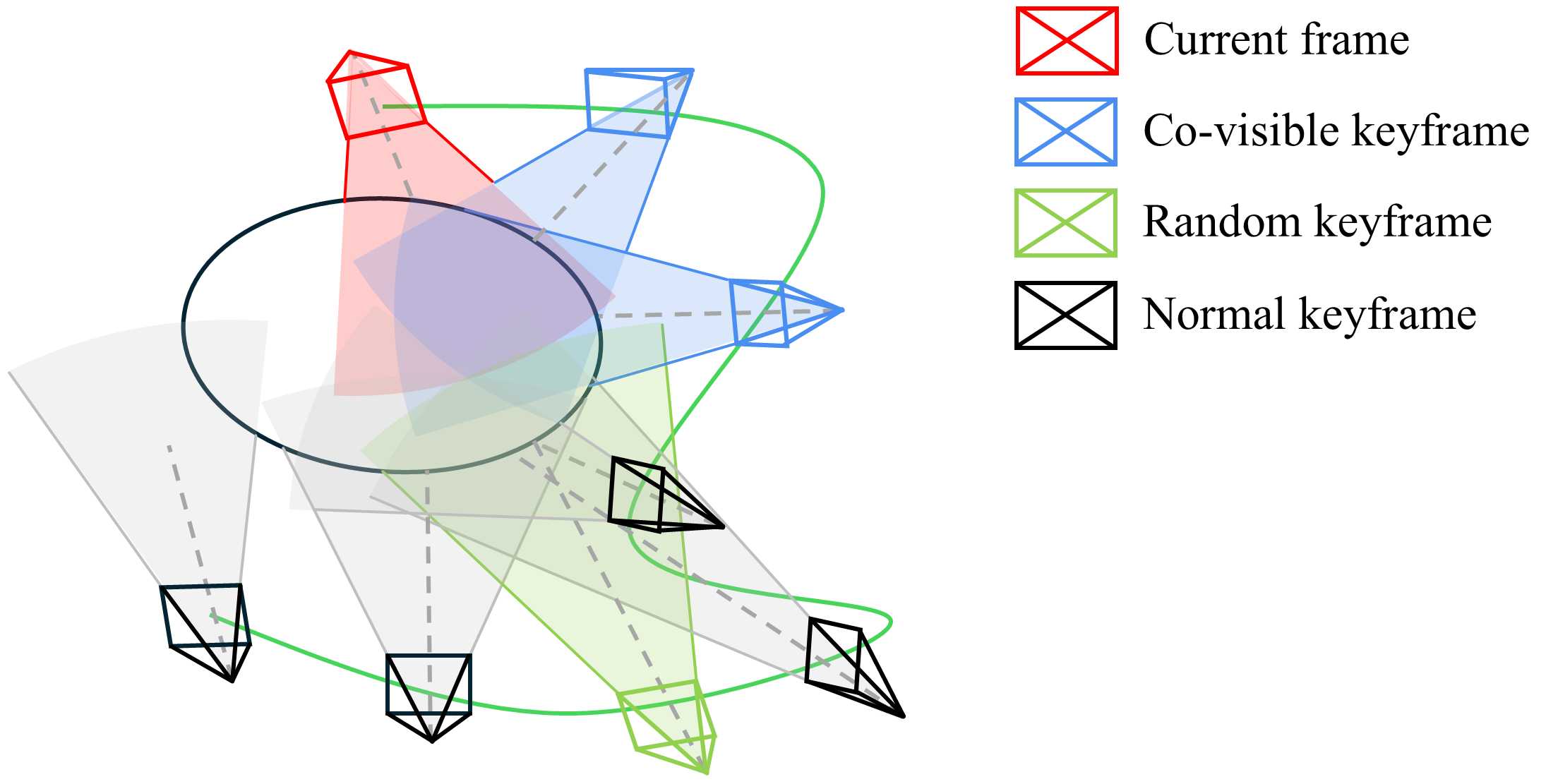}
	\caption{Illustration of Keyframe Selection Strategy. The red frustum and its sector represent the current frame's observed field of view. The blue frustums indicate co-visibility keyframes with a partially overlapping field of view with the current frame. The black frustums represent normal keyframes, while the green frustums are randomly selected keyframes from other frames.}
	\label{fig3}
\end{figure}

To further optimize the map, additional filtering is applied to the keyframes. As shown in Fig. \ref{fig3}, we select keyframes with a co-visibility greater than 70\% with the current frame as co-visible keyframes. Additionally, to mitigate the effects of scene forgetting and gaussian artifacts, we randomly sample 30\% of the remaining keyframes as random keyframes.

\subsection{Map Optimization}
Our system optimizes both sparse and dense gaussian maps through local and global optimization. The co-visible keyframes are used for local optimization of the neighboring region observed in the current frame, while global optimization is achieved jointly by the co-visible keyframes and randomly selected keyframes.

To ensure stability and avoid uncontrolled gaussian scaling, we introduce a regularization loss $\mathcal{L}_{reg}$:
\begin{equation}
	\label{eq15}
	\begin{aligned}\mathcal{L}_{reg}=\frac{1}{n}\sum_{i=1}^{n}|S_{i,1:2}-\bar{S}_{1:2}|+\frac{1}{n}\sum_{i=1}^{n}\left|S_{i,3}-\varepsilon\right|\end{aligned},
\end{equation}
where $S\in\mathbb{R}^{n\times3}$ represents the gaussian scale parameter matrix, where $n$ is the sample count, and $\varepsilon \to 0$. $\bar{S}_{1:2}$ denotes the mean of the first two scale parameters. This regularization loss $\mathcal{L}_{reg}$ encourages consistency of gaussian scales across dimensions and promotes flattened alignment along the surface normal direction in the neighborhood.

The mapping loss combines color loss, depth loss, structural similarity (SSIM) loss, and regularization loss:
\begin{equation}
	\label{eq16}
	\begin{aligned}
		\mathcal{L}{o}{s}{s}= 
		&\lambda_i||I-I_{gt}||_1 + \lambda_d||D-D_{gt}||_1\\
		& + (1-\lambda_i)(1-ssim(I,I_{gt})) + \lambda_r\mathcal{L}_{reg}
	\end{aligned},
\end{equation}
where $\lambda_i$ and $\lambda_d$ represent the weights for color loss and depth loss, respectively. $I$ and $I_{gt}$ denote the RGB rendered image and the ground truth image, respectively. $D$ and $D_{gt}$ represent the depth rendered image and the ground truth depth image. $ssim(\cdot)$ denotes the structural similarity calculation, and $\lambda_r$ is the weight for the regularization loss.

Mapping optimization and tracking run in parallel threads, significantly enhancing speed and ensuring real-time SLAM performance. The dense map updates on tracking keyframes, and both sparse and dense maps share computational resources, leading to unified optimization without additional computational cost.

\begin{figure*}[!t]
	\centering
	\includegraphics[width=0.85\textwidth]{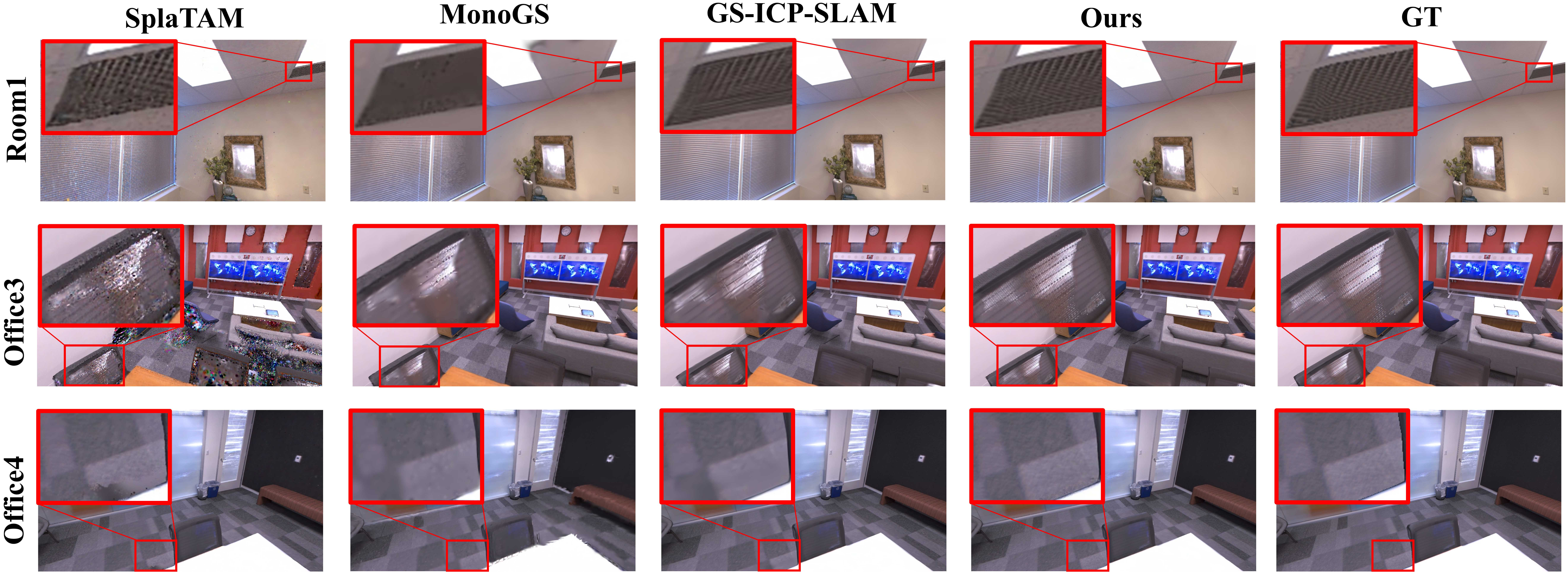}
	\caption{Qualitative result comparison on the Replica dataset. Detail zoom-ins from three scenes are presented. Our method outperforms other frameworks in the reconstruction of map details.}
	\label{fig4}
\end{figure*}

\begin{figure}[!t]
	\centering
	\includegraphics[width=0.45\textwidth]{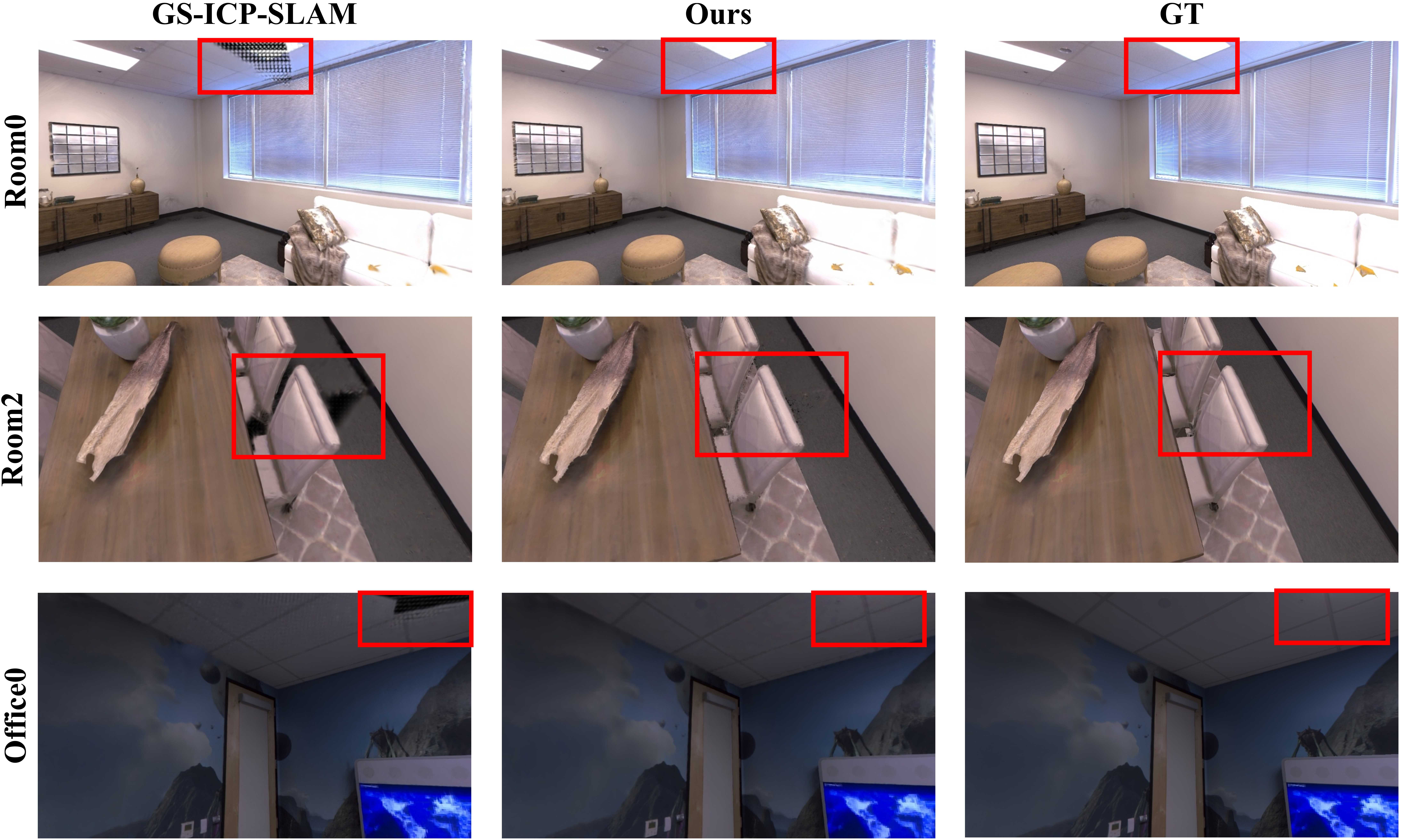}
	\caption{Performance Comparison of gaussian missing region check strategies. Our method effectively and promptly fills in gaussians in regions observed in very few frames.}
	\label{fig5}
\end{figure}

\section{Experiments}
\subsection{Experimental Setup}

\textbf{Datasets.} We evaluate our method using the Replica and TUM RGB-D datasets. The Replica dataset provides synthetic data with precise RGB images and depth maps. TUM RGB-D dataset have relatively lower image quality, requiring greater robustness and accuracy in dense mapping and tracking methods.

\textbf{Implementation Details.} Our experiments were conducted on an NVIDIA GeForce RTX 4090 24GB GPU. The mapping optimization and tracking processes run in parallel threads, sharing gaussian data across threads for efficiency.

\textbf{Evaluation Metrics.} For map reconstruction performance, we assess rendering quality using PSNR, SSIM, and LPIPS metrics. For camera tracking accuracy, we use the Root Mean Square Error of Absolute Trajectory Error (ATE RMSE) as the main evaluation metric. For evaluating system real-time performance, we use the overall system frame rate (FPS) rather than the rendering speed or the speed of individual components.

\textbf{Baseline Methods.} We compare our approach with NeRF-based neural implicit SLAM methods, including NICE-SLAM \cite{zhu2022nice}, Point-SLAM \cite{sandstrom2023point}, and Co-SLAM \cite{wang2023co}. Additionally, we compare with state-of-the-art 3DGS-based explicit SLAM methods like GS-SLAM \cite{yan2024gs}, SplaTAM \cite{keetha2024splatam}, MonoGS \cite{matsuki2024gaussian}, CG-SLAM \cite{hu2025cg}, and GS-ICP-SLAM \cite{ha2024rgbd}.

\subsection{Camera Tracking Accuracy Evaluation}

\begin{table}[]
	\centering
	\caption{Camera Pose Estimation Results on the Replica Dataset (ATE RMSE$\downarrow$[cm]). Our method demonstrates superior performance across all 8 scenes, outperforming the current state-of-the-art (SOTA) baselines. Some baseline data is sourced from \cite{hu2025cg}.}
	\label{tab1}
	\resizebox{0.47\textwidth}{!}{%
		\begin{tabular}{lccccccccc}
			\toprule
			Method & R0   & R1   & R2   & OF0  & OF1  & OF2  & OF3  & OF4  & Avg. \\ \midrule
			NICE-SLAM \cite{zhu2022nice}     & 0.97 & 1.31 & 1.07 & 0.88 & 1.00 & 1.06 & 1.10 & 1.13 & 1.06 \\
			Point-SLAM \cite{sandstrom2023point}   & 0.56 & 0.47 & 0.30 & \cellcolor{yellow!20}0.35 & 0.62 & 0.55 & 0.72 & 0.73 & 0.54 \\
			Co-SLAM \cite{wang2023co}      & 0.77 & 1.04 & 1.09 & 0.58 & 0.53 & 2.05 & 1.49 & 0.84 & 0.99 \\
			GS-SLAM \cite{yan2024gs}      & 0.48 & 0.53 & 0.33 & 0.52 & 0.41 & 0.59 & 0.46 & 0.70 & 0.50 \\
			SplaTAM \cite{keetha2024splatam}\textsuperscript{$\dagger$}       & \cellcolor{orange!20}0.27 & 0.31 & 0.63 & 0.49 & \cellcolor{yellow!20}0.22 & 0.30 & 0.35 & \cellcolor{yellow!20}0.52 & \cellcolor{yellow!20}0.39 \\
			MonoGS (RGB-D) \cite{matsuki2024gaussian}\textsuperscript{$\dagger$} & 0.35 & \cellcolor{orange!20}0.26 & \cellcolor{yellow!20}0.27 & 0.41 & 0.40 & \cellcolor{orange!20}0.22 & \cellcolor{red!20}0.14 & 2.10 & 0.52 \\
			CG-SLAM \cite{hu2025cg}      & \cellcolor{yellow!20}0.29 & \cellcolor{yellow!20}0.27 & \cellcolor{orange!20}0.25 & \cellcolor{orange!20}0.33 & \cellcolor{orange!20}0.14 & \cellcolor{yellow!20}0.28 & \cellcolor{yellow!20}0.31 & \cellcolor{orange!20}0.29 & \cellcolor{orange!20}0.27 \\
			\textbf{Ours}          & \cellcolor{red!20} 0.14 & \cellcolor{red!20}0.17 & \cellcolor{red!20}0.10 & \cellcolor{red!20}0.16 & \cellcolor{red!20}0.13 & \cellcolor{red!20}0.16 & \cellcolor{orange!20}0.16 & \cellcolor{red!20}0.20  & \cellcolor{red!20}0.15 \\ \bottomrule
		\end{tabular}%
	}
	\begin{tablenotes}
		\footnotesize
		\item \textsuperscript{$\dagger$} denotes the reproduced results by running officially released code.
		\vspace*{-\baselineskip}
	\end{tablenotes}
\end{table}

The tables in this paper contains three colors, representing the \colorbox{red!20}{best}, \colorbox{orange!20}{second best}, and \colorbox{yellow!20}{third best}, respectively. As shown in Table \ref{tab1}, our method demonstrates excellent tracking performance across eight scenes in the Replica dataset. This success is due to the shared gaussian distribution assumptions between GICP and our map's gaussian representation. GICP achieves tracking by leveraging the 3D gaussian distribution within the sparse map. In contrast, other baseline methods typically track camera poses by minimizing 2D image rendering differences with ground truth images, which demands maintaining dense maps. The upkeep of dense maps is computationally expensive, and their quality greatly affects tracking accuracy, as mapping errors can propagate into tracking. Our method, by maintaining a lightweight sparse map for GICP tracking, avoids the need to project 3D maps into 2D, achieving direct tracking within the 3D space.

In Table \ref{tab2}, the tracking performance of various methods on the TUM RGB-D dataset is presented. Due to lower image quality and motion blur, tracking accuracy on the TUM dataset is generally reduced compared to Replica. Our method achieves tracking performance comparable to advanced NeRF-based SLAM and 3DGS-based SLAM approaches but offers a significantly faster system speed, with at least an order of magnitude advantage. While both GS-ICP-SLAM \cite{ha2024rgbd} and our method operate in real time, our method achieves higher tracking accuracy.

\begin{table}[]
	\centering
	\caption{Camera Pose Estimation Results on the TUM-RGBD Dataset (ATE RMSE$\downarrow$[cm]). Our method achieves tracking accuracy on par with the state-of-the-art (SOTA) in this dataset, with the system running at a frame rate (FPS$\uparrow$) that demonstrates advanced performance. Some baseline data is sourced from \cite{ha2024rgbd, hu2025cg}.}
	\label{tab2}
	\resizebox{0.47\textwidth}{!}{%
		\begin{tabular}{lccccc}
			\toprule
			Method & fr1/desk & fr2/xyz & fr3/office & Avg. & FPS$\uparrow$   \\ \midrule
			NICE-SLAM \cite{zhu2022nice}    & 2.8      & 2.1     & 7.2        & 4.0  & 0.08  \\
			Point-SLAM \cite{sandstrom2023point}   & 2.7      & \cellcolor{orange!20}1.3     & 3.9        & 2.6  & 0.22  \\
			Co-SLAM \cite{wang2023co}      & 2.7      & 1.9     & 2.6        & \cellcolor{yellow!20}2.4  & -     \\
			GS-SLAM \cite{yan2024gs}      & 3.3      & \cellcolor{orange!20}1.3     & 6.6        & 3.7  & -     \\
			SplaTAM \cite{keetha2024splatam}\textsuperscript{$\dagger$}       & 3.3      & \cellcolor{orange!20}1.3     & 5.1        & 3.2  & 0.22  \\
			MonoGS (RGB-D) \cite{matsuki2024gaussian}\textsuperscript{$\dagger$} & \cellcolor{red!20}1.5      & 1.6     & \cellcolor{red!20}1.7        & \cellcolor{red!20}1.6  & \cellcolor{yellow!20}1.64  \\
			CG-SLAM \cite{hu2025cg}      & \cellcolor{orange!20}2.4      & \cellcolor{red!20}1.2     & \cellcolor{yellow!20}2.5        & \cellcolor{orange!20}2.0  & -     \\
			GS-ICP-SLAM \cite{ha2024rgbd}\textsuperscript{$\dagger$}   & 2.7      & 1.8     & 2.7        & \cellcolor{yellow!20}2.4  & \cellcolor{orange!20}29.96 \\
			\textbf{Ours}          & \cellcolor{yellow!20}2.5      & \cellcolor{yellow!20}1.5     & \cellcolor{orange!20}2.1        & \cellcolor{orange!20}2.0  & \cellcolor{red!20}44.09 \\ \bottomrule
		\end{tabular}%
	}
\end{table}

\subsection{Mapping Quality Evaluation}
Table \ref{tab3} presents our method's mapping quality and system speed. Our approach achieves state-of-the-art or near-optimal performance in mapping quality across most metrics while maintaining a speed advantage of one to two orders of magnitude. This efficiency is due to our shared sparse and dense maps, where sparse map tracking and shared gaussians conserve computational resources, and frequency-domain-informed adaptive gaussian densification ensures high map quality.

\begin{table}[]
	\centering
	\caption{Rendering Results on the Replica Dataset. Our method achieves an optimal balance between system speed and mapping quality. Some baseline data is sourced from \cite{sandstrom2023point, ha2024rgbd}.}
	\label{tab3}
	\resizebox{0.47\textwidth}{!}{%
		\begin{tabular}{l|l|cccccccccc}
			\toprule
			Method & Metrics      & R0     & R1     & R2     & OF0   & OF1   & OF2   & OF3   & OF4   & Avg.      & FPS $\uparrow$  \\ \midrule
			NICE-SLAM \cite{zhu2022nice}    & PSNR{[}dB{]}$\uparrow$ & 22.12  & 22.47  & 24.52  & 29.07 & 30.34 & 19.66 & 22.23 & 24.94 & 24.42     &     \\
			& SSIM $\uparrow$        & 0.689  & 0.757  & 0.814  & 0.874 & 0.886 & 0.797 & 0.801 & 0.856 & 0.809     &      -\\
			& LPIPS  $\downarrow$      & 0.330   & 0.271  & 0.208  & 0.229 & 0.181 & 0.235 & 0.209 & 0.198 & 0.233     &      \\ \midrule
			Point-SLAM \cite{sandstrom2023point}    & PSNR{[}dB{]}$\uparrow$ & \cellcolor{yellow!20}33.38  & 34.10   & 36.32  & 38.72 & 39.31 & 34.22 & 34.10  & 34.82 & 35.62     &   \\
			& SSIM $\uparrow$        & \cellcolor{red!20}0.979  & \cellcolor{red!20}0.977  & \cellcolor{red!20}0.985  & \cellcolor{orange!20}0.985 & \cellcolor{orange!20}0.987 & 0.962 & 0.963 & \cellcolor{red!20}0.981 & \cellcolor{red!20}0.977     &   0.3   \\
			& LPIPS $\downarrow$       & 0.097  & 0.115  & 0.101  & 0.089 & 0.110  & 0.152 & 0.119 & 0.131 & 0.114     &      \\ \midrule
			GS-SLAM \cite{yan2024gs}      & PSNR{[}dB{]}$\uparrow$ & 31.56  & 32.86  & 32.59  & 38.70  & 41.17 & 32.36 & 32.03 & 32.92 & 34.27     &  \\
			& SSIM  $\uparrow$       & \cellcolor{orange!20}0.968  & \cellcolor{orange!20}0.973  & 0.971  & \cellcolor{red!20}0.986 & \cellcolor{red!20}0.993 & \cellcolor{red!20}0.978 & \cellcolor{red!20}0.970  & 0.968 & \cellcolor{orange!20}0.975     &  \cellcolor{yellow!20}8.34    \\
			& LPIPS $\downarrow$       & 0.094  & \cellcolor{yellow!20}0.075  & 0.093  & 0.050  & \cellcolor{red!20}0.033 & 0.094 & 0.110  & \cellcolor{yellow!20}0.112 & 0.082     &      \\ \midrule
			SplaTAM \cite{keetha2024splatam}\textsuperscript{$\dagger$}       & PSNR{[}dB{]}$\uparrow$ & 32.60   & 33.63  & 34.91  & 38.15 & 39.05 & 31.89 & 30.18 & 32.01 & 34.05   &  \\
			& SSIM $\uparrow$        & \cellcolor{yellow!20}0.975  & 0.969  & \cellcolor{orange!20}0.982  & 0.981 & 0.981 & 0.966 & 0.951 & 0.948 & 0.969  &   0.18   \\
			& LPIPS $\downarrow$       & \cellcolor{yellow!20}0.070   & 0.097  & \cellcolor{yellow!20}0.073  & 0.088 & 0.094 & 0.100   & 0.118 & 0.154 & 0.099   &      \\ \midrule
			MonoGS (RGB-D) \cite{matsuki2024gaussian}\textsuperscript{$\dagger$} & PSNR{[}dB{]}$\uparrow$ & 33.21 & \cellcolor{yellow!20}35.88 & \cellcolor{yellow!20}36.86 & \cellcolor{yellow!20}40.49 & \cellcolor{yellow!20}41.39 & \cellcolor{yellow!20}35.62 & \cellcolor{yellow!20}35.48 & \cellcolor{yellow!20}33.65 & \cellcolor{yellow!20}36.57 &  \\
			& SSIM $\uparrow$        & 0.937  & 0.954  & 0.961  & 0.974 & 0.975 & 0.958 & 0.957 & 0.940  & 0.957     &    0.81  \\
			& LPIPS $\downarrow$       & 0.081  & 0.092  & 0.075  & 0.061 & 0.053 & \cellcolor{yellow!20}0.071 & \cellcolor{yellow!20}0.059 & \cellcolor{yellow!20}0.112 & \cellcolor{yellow!20}0.076    &      \\ \midrule
			GS-ICP-SLAM \cite{ha2024rgbd}\textsuperscript{$\dagger$}   & PSNR{[}dB{]}$\uparrow$ & \cellcolor{orange!20}35.11  & \cellcolor{orange!20}37.28  & \cellcolor{orange!20}38.11  & \cellcolor{orange!20}42.38 & \cellcolor{orange!20}42.76 & \cellcolor{red!20}36.77 & \cellcolor{orange!20}36.80 & \cellcolor{orange!20}38.54 & \cellcolor{orange!20}38.55  &    \\
			& SSIM $\uparrow$        & 0.960   & 0.968  & 0.973  & \cellcolor{yellow!20}0.984 & 0.982 & \cellcolor{yellow!20}0.971 & \cellcolor{yellow!20}0.968 & \cellcolor{yellow!20}0.967 & 0.970   &   \cellcolor{orange!20}29.95   \\
			& LPIPS $\downarrow$       & \cellcolor{orange!20}0.053  & \cellcolor{orange!20}0.051  & \cellcolor{orange!20}0.053  & \cellcolor{orange!20}0.032  & \cellcolor{yellow!20}0.036 & \cellcolor{orange!20}0.048 & \cellcolor{orange!20}0.047 & \cellcolor{orange!20}0.049 & \cellcolor{orange!20}0.045  &      \\ \midrule
			\textbf{Ours}          & PSNR{[}dB{]}$\uparrow$ & \cellcolor{red!20}35.27  & \cellcolor{red!20}38.05  & \cellcolor{red!20}38.63  & \cellcolor{red!20}42.73 & \cellcolor{red!20}43.18 & \cellcolor{orange!20}36.42 & \cellcolor{red!20}37.04 & \cellcolor{red!20}38.66 & \cellcolor{red!20}38.75     &    \\
			& SSIM $\uparrow$        & 0.961  & \cellcolor{yellow!20}0.972  & \cellcolor{yellow!20}0.975  & \cellcolor{yellow!20}0.984 & \cellcolor{yellow!20}0.984 & \cellcolor{orange!20}0.973 & \cellcolor{orange!20}0.969 & \cellcolor{orange!20}0.972 & \cellcolor{yellow!20}0.974   &    \cellcolor{red!20}32.75  \\
			& LPIPS $\downarrow$       & \cellcolor{red!20}0.045  & \cellcolor{red!20}0.043   & \cellcolor{red!20}0.045  & \cellcolor{red!20}0.028 & \cellcolor{orange!20}0.035  & \cellcolor{red!20}0.045 & \cellcolor{red!20}0.040  & \cellcolor{red!20}0.046 & \cellcolor{red!20}0.041  &      \\ \bottomrule
		\end{tabular}%
	}
\end{table}

Fig. \ref{fig4} presents a qualitative comparison of results, using three scenes as examples to demonstrate the high demands for fine-grained mapping in the system. In the zoomed-in view of the iron mesh detail in Room1, the SplaTAM result shows chaotic textures, MonoGS fails to reconstruct the iron mesh texture, and GS-ICP-SLAM only reconstructs part of the texture. In Office3, the chair backrest is semi-transparent and consists of rows of black dots forming lines. GS-ICP-SLAM fails to reconstruct the black dots and instead reconstructs them as black lines. For the floor texture details in Office4, our method outperforms the others. In contrast, our method captures the fine texture differences in these scenes, with minimal deviation from the ground truth. The successful reconstruction of these details is attributed to our Fourier frequency segmentation mechanism, which initializes a dense distribution of small gaussians at high-frequency locations. 

We employ a gaussian missing region check strategy for each new keyframe. Fig. \ref{fig5} provides an intuitive comparison between our strategy and GS-ICP-SLAM. GS-ICP-SLAM adds new gaussians by populating new keyframes with uniformly and equidistantly sampled sparse gaussian points. While this strategy is simple and efficient to execute, it leads to insufficient gaussian representation in regions observed in few frames. This issue can result in holes in the map. Our deficient region inpainting strategy can promptly fill these holes and correct regions with significant map discrepancies.

As shown in the Table \ref{tab4}, compared to other methods, our system has the shortest tracking time, and the single mapping time is the lowest. Surprisingly, the iteration count is fewer than three, which is 5 to 100 times fewer than the mapping iteration count of other methods. This demonstrates that the frequency-domain guided SLAM method proposed in this paper can converge quickly with very few iterations.

\begin{figure}[!t]
	\centering
	\includegraphics[width=0.42\textwidth]{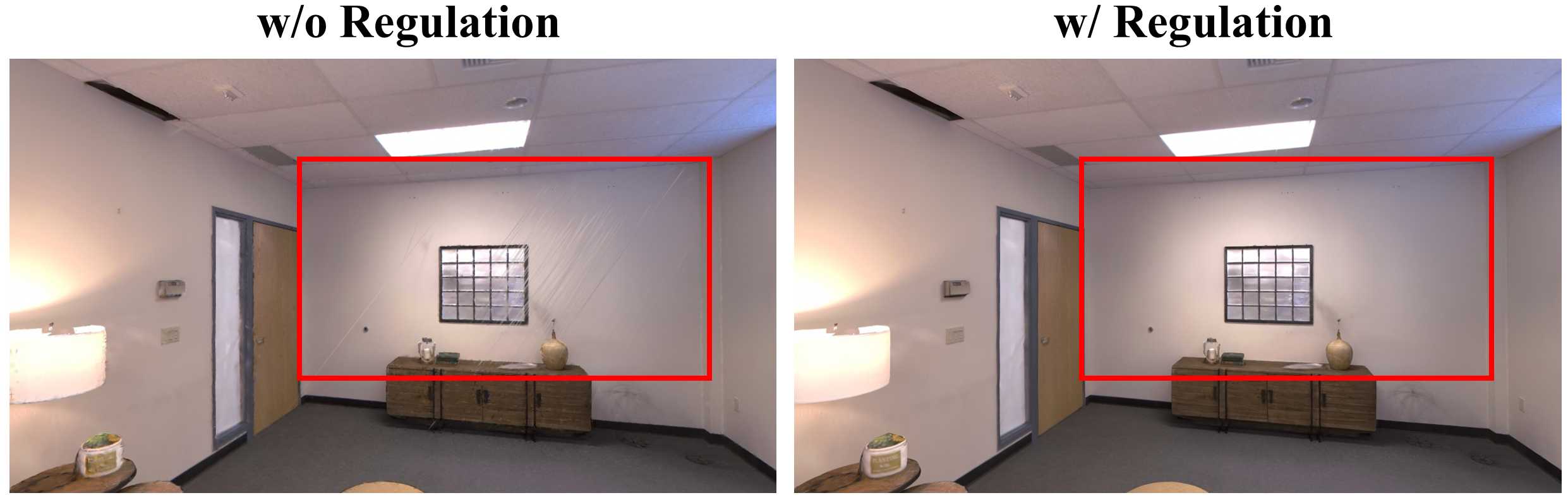}
	\caption{Ablation comparison of gaussian regularization loss.}
	\label{fig6}
\end{figure}

\subsection{Ablation Study}

\textbf{Keyframe Selection.} We evaluated the impact of different keyframe selection strategies on mapping performance using eight Replica dataset scenes. The first approach randomly selects keyframes, the second optimizes only co-visible keyframes, and the third combines both co-visible and randomly selected keyframes. Table \ref{tab5} shows that optimizing only random keyframes reduces optimization opportunities in recently observed areas, while focusing solely on co-visible keyframes can over-stretch certain gaussians and lead to unobserved scene regions being forgotten. Combining both strategies provides optimal results by balancing recent and global scene coverage.

\begin{table}[]
	\centering
	\caption{Execution time of each module in the system on Replica / Office0. {[}ms $\times$ it{]} denotes the single execution time and the number of iterations.  The baseline data is sourced from \cite{hu2025cg}.}
	\label{tab4}
	\resizebox{0.35\textwidth}{!}{%
		\begin{tabular}{lccc}
			\toprule
			\multirow{2}{*}{Method}    & Tracking & Mapping  & System \\
			& {[}ms $\times$ it{]$\downarrow$}& {[}ms $\times$ it{]$\downarrow$}& FPS $\uparrow$ \\  \midrule
			Vox-Fusion \cite{yang2022vox}     & 23.61 $\times$ 30       & 86.55 $\times$ 10 & 1.1 \\
			NICE-SLAM \cite{zhu2022nice} & \cellcolor{yellow!20}6.19 $\times$ 10        & 91.59 $\times$ 60 & 0.98 \\
			Co-SLAM \cite{wang2023co} & \cellcolor{orange!20}4.45 $\times$ 10        & \cellcolor{orange!20}10.9 $\times$ 10 & \cellcolor{orange!20}14.2 \\
			Point-SLAM \cite{sandstrom2023point} & 6.14 $\times$ 40        & 22.25 $\times$ 300 & 0.48 \\
			GS-SLAM \cite{yan2024gs} & 11.9 $\times$ 10        & 12.8 $\times$ 100 & 8.34 \\
			SplaTAM \cite{keetha2024splatam} & 41.7 $\times$ 40        & 50.1 $\times$ 60 & 0.21 \\
			CG-SLAM \cite{hu2025cg} & 7.89 $\times$ 15        & \cellcolor{yellow!20}12.2 $\times$ 60 & \cellcolor{yellow!20}8.5 \\
			\textbf{Ours}                                 & \cellcolor{red!20}29.8 $\times$ \textbf{1}        & \cellcolor{red!20}10.7 $\times$ \textbf{2.8} & \cellcolor{red!20} 33.04 \\ \bottomrule
		\end{tabular}%
	}
\end{table}

\begin{table}[]
	\centering
	\caption{Ablation study of keyframe selection on the Replica dataset. The results represent the average across 8 scenes.}
	\label{tab5}
	\resizebox{0.35\textwidth}{!}{%
		\begin{tabular}{cccc}
			\toprule
			Method                               & PSNR{[}dB{]$\uparrow$} & SSIM $\uparrow$  & LPIPS $\downarrow$ \\ \midrule
			w/o co-visible keyframes     & \cellcolor{orange!20}37.67        & \cellcolor{orange!20}0.968 & \cellcolor{orange!20}0.049 \\
			w/o random keyframes & \cellcolor{yellow!20}34.55        & \cellcolor{yellow!20}0.949 & \cellcolor{yellow!20}0.077 \\
			\textbf{Ours}                                 & \cellcolor{red!20}38.75        & \cellcolor{red!20}0.974 & \cellcolor{red!20}0.041 \\ \bottomrule
		\end{tabular}%
	}
\end{table}

\begin{table}[]
	\centering
	\caption{Ablation study of adaptive gaussian densification on the Replica / Office0 dataset.}
	\label{tab6}
	\resizebox{0.47\textwidth}{!}{%
		\begin{tabular}{cccccc}
			\toprule
			\multicolumn{2}{c}{Method}                            & PSNR{[}dB{]$\uparrow$}  & SSIM $\uparrow$  & LPIPS $\downarrow$ & Memory Usage$\downarrow$ \\ \midrule
			\multirow{2}{*}{Sparse equidistant} & Large gaussians & 39.78 & 0.973 & 0.064 & \cellcolor{red!20}29.6 M        \\
			& Small gaussians & \cellcolor{yellow!20}41.94  & \cellcolor{yellow!20}0.981 & \cellcolor{yellow!20}0.037 & \cellcolor{orange!20}54.4 M        \\ \midrule
			\multirow{2}{*}{Dense equidistant}  & Large gaussians & 40.62 & 0.978 & 0.044 & 112.4M       \\ 
			& Small gaussians & \cellcolor{orange!20}42.68 & \cellcolor{red!20}0.984 & \cellcolor{red!20}0.022  & 157.4 M       \\ \midrule
			\multicolumn{2}{c}{\textbf{Adaptive gaussian densification}}   & \cellcolor{red!20}42.73 & \cellcolor{red!20}0.984 & \cellcolor{orange!20}0.028 & \cellcolor{yellow!20}68.1 M        \\ \bottomrule
		\end{tabular}%
	}
\end{table}

\textbf{Adaptive Gaussian Densification.} To assess the effect of adaptive densification based on frequency-domain analysis, we conducted ablation studies varying gaussian density and radius. Table \ref{tab6} shows results in Replica Office0, where equidistant sparse mapping yields lower quality than equidistant dense and adaptive dense methods. However, sparse mapping uses the fewest gaussians and has lower memory requirements. Large-radius gaussians yield weaker maps than small-radius gaussians, suggesting the latter's advantage in capturing details. However, small-radius gaussians in equidistant dense initialization lead to increased gaussian count and memory use. Our adaptive densification balances mapping quality and memory usage by applying sparse large-radius gaussians in low-frequency regions and dense small-radius gaussians in high-frequency areas.

\textbf{Regularization.} As shown in Fig. \ref{fig6}, adding gaussian regularization loss improves the mapping quality. This is because the regularization constrains the gaussian distribution, preventing it from being excessively elongated in any particular direction. Experimental results demonstrate that the regularization constraint leads to higher mapping quality.

\section{Conclusion}
We introduce a novel SLAM system based on frequency-domain analysis that initializes gaussians according to high- and low-frequency regions, resulting in detailed mapping in high-frequency areas. By combining a gaussian missing region check strategy, the system effectively avoids incomplete reconstructions in regions visible in few frames. The proposed shared-resource mechanism for dense and sparse gaussian maps significantly enhances system speed. This framework achieves state-of-the-art mapping quality while maintaining real-time operational speed. The proposed method opens up a new research avenue for analyzing 3DGS SLAM from the frequency domain perspective. We believe that the rich information in the frequency domain will drive the further development of SLAM.

\bibliographystyle{IEEEtran}
\bibliography{IEEEexample}

\end{document}